\documentclass[10pt,twocolumn,letterpaper]{article}

\usepackage{cvpr}
\usepackage{times}
\usepackage{epsfig}
\usepackage{graphicx}
\usepackage{amsmath}
\usepackage{amssymb}
\usepackage{algorithm}
\usepackage{algorithmic}

\usepackage[breaklinks=true,bookmarks=false]{hyperref}

\cvprfinalcopy 

\def\cvprPaperID{****} 

\ifcvprfinal\fi
\begin{document}

\title{Physics Inspired Optimization on Semantic Transfer Features: An Alternative Method for Room Layout Estimation}

\author{Hao Zhao$^1$\thanks{This work was done when Hao Zhao was an intern at Intel Labs China
supervised by Anbang Yao who is responsible for correspondence.}, Ming Lu$^1$, Anbang Yao$^2$, Yiwen Guo$^2$, Yurong Chen$^2$, Li Zhang$^1$ \\
$^1$Department of Electronic Engineering, Tsinghua University \\
$^2$Cognitive Computing Laboratory, Intel Labs China \\
{\tt\small \{zhao-h13@mails,lu-m13@mails,chinazhangli@mail\}.tsighua.edu.cn} \\
{\tt\small \{anbang.yao, yiwen.guo, yurong.chen\}@intel.com}}

\maketitle
\thispagestyle{empty}

\begin{abstract}
   In this paper, we propose an alternative method to estimate room layouts of cluttered indoor scenes. This method enjoys the benefits of two novel techniques. The first one is \textbf{semantic transfer} (ST), which is: (1) a formulation to integrate the relationship between scene clutter and room layout into convolutional neural networks; (2) an architecture that can be end-to-end trained; (3) a practical strategy to initialize weights for very deep networks under unbalanced training data distribution. ST allows us to extract highly robust features under various circumstances, and in order to address the computation redundance hidden in these features we develop a principled and efficient inference scheme named \textbf{physics inspired optimization} (PIO). PIO's basic idea is to formulate some phenomena observed in ST features into mechanics concepts. Evaluations on public datasets LSUN and Hedau show that the proposed method is more accurate than state-of-the-art methods.
%
\end{abstract}

\section{Introduction}

Given an input RGB image, a room layout estimation algorithm should output all the wall-floor, wall-wall, and wall-ceiling edges (depicted by Fig~\ref{fig:problem}). This is a fundamental indoor scene understanding task as it can provide a strong prior for other tasks like depth recovery from a single RGB image \cite{eigen2014depth}\cite{eigen2015predicting} or indoor object pose estimation \cite{song2014sliding}\cite{gupta2015aligning}\cite{song2016deep}. Besides, the room layout itself provides a high-level representation of an indoor scene for emerging applications like intelligent robots and augmented reality. This problem draws constant attention since the publication of the seminal work \cite{hedau2009recovering}, and there are two lines of followers:

\begin{figure}
\begin{center}
\includegraphics[width=8.5cm]{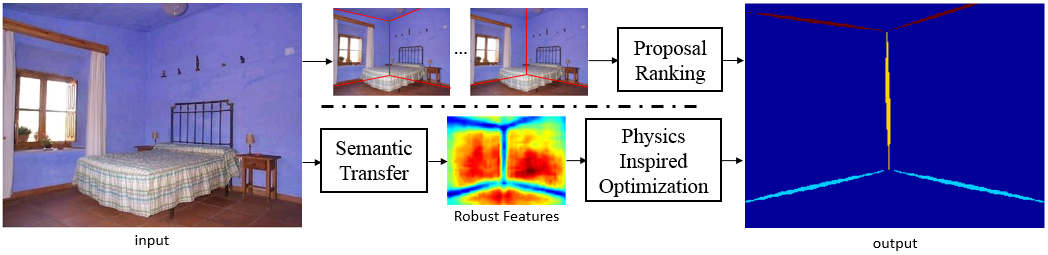}
\end{center}
   \caption{Above is the overview of conventional methods. Below is the overview of our method. Better viewed electronically.}
\label{fig:problem}
\end{figure}

 (1) As the upper part of Fig~\ref{fig:problem} shows, conventional methods follow a \emph{proposing-ranking} scheme. Typically, the \emph{proposing} part consists of three sub-modules as edge detection, vanishing point voting and ray sampling. With hand-crafted features and structured inference techniques, the \emph{ranking} part outputs the best layout proposal, sometimes along with a representation of the clutter.

 (2) Recent methods \cite{mallya2015learning}\cite{dasgupta2016delay}\cite{ren2016cfile} achieve dramatic performance improvements via features produced by fully convolutional networks (FCNs). \cite{mallya2015learning}\cite{ren2016cfile} still follow the traditional \emph{proposing-ranking} scheme. \cite{dasgupta2016delay} is a proposal-free solution in which all those steps about proposal generation are eliminated. And instead of proposal ranking, in \cite{dasgupta2016delay} inference is achieved through an optimization module.

 Alternative to these two lines of works, we propose a method that features the advantages of both yet goes beyond them. It is illustrated by the lower part of Fig~\ref{fig:problem} and the motivations are in two folds:

\textbf{Conventional methods.} They provide many useful insights about indoor scene understanding. \cite{hedau2009recovering} and its followers \cite{wang2010discriminative}\cite{schwing2012efficient}\cite{schwing2013box}\cite{del2012bayesian}\cite{del2013understanding}\cite{zhao2013scene} explore different ways to model the relationship between room layout and scene clutter. This effort is reasonable because the major challenges of room layout estimation lie here. Take Fig~\ref{fig:problem} for example, over 50\% of wall-floor edge pixels is occluded by the bed. If the bed does not exist, this task will become much easier. However, these insights are not visited in recent FCN-based room layout estimation works. When designing networks, they treat FCNs as black boxes, taking no scene clutter information into consideration. As modelling meaningful concepts with a neural network has always been difficult, it motivates us to explore the possibility to describe scene clutter within an FCN.

\textbf{FCN-based methods.} Unlike \cite{mallya2015learning}\cite{ren2016cfile} which still follow the \emph{proposing-ranking} scheme, \cite{dasgupta2016delay}'s framework shows intriguing compactness. However, its optimization is primitive in which the sampled solution space is exhaustively searched and no gradient is modeled. So the second motivation of this paper is to develop a principled, gradient-based, and efficient optimization algorithm for this task.

Guided by the first motivation, we propose \textbf{Semantic transfer} (ST) which has three features from three different perspectives: 1) As a discriminative model, it integrates the relationship between room layout and scene clutter into an FCN. 2) As an architecture, it enjoys the benefit of end-to-end training. 3) As a training strategy, it provides better network initialization and allows us to train a very deep network under unbalanced training data distribution. ST provides highly robust features under various circumstances. Accordingly we propose an inference technique named \textbf{Physics inspired optimization} (PIO). ST and PIO play different yet closely interdependent roles because the core idea of PIO is to formulate some phenomena observed in ST feature maps with mechanics concepts.

\section{Related Works}

\textbf{Conventional methods.} The standard definition of room layout estimation is firstly introduced by \cite{hedau2009recovering}. It clusters edges into lines joining at three vanishing points, according to the famous Manhattan assumption \cite{coughlan1999manhattan}. Then a lot of layout proposals are generated by ray sampling. Hand-crafted features are used to learn a regressor for proposal ranking. Later on, many works try to improve this framework. \cite{ramalingam2013manhattan} detects conjunctions instead of edges and modifies proposal generation and ranking accordingly. While ranking room layouts, \cite{wang2010discriminative} simultaneously estimates a clutter mask. \cite{schwing2012efficient} aims to improve the inference efficiency of methods like \cite{wang2010discriminative}. Going beyond estimating clutter mask, \cite{schwing2013box} estimates objects' 3D bounding boxes and room layout during inference. Except for learnt clutter representations, \cite{del2012bayesian} incorporates furniture shape prior. In \cite{del2013understanding}'s formulation, furniture is modeled with parts instead of a box. \cite{zhao2013scene} goes even further by modelling furniture relationship with scene grammars.

\textbf{FCN-based methods.} Recently \cite{mallya2015learning} trains an FCN for pixel-wise edge labelling, with every pixel assigned a label from this 4-class set $S$ \{background (bg), wall-floor edge (wf), wall-wall edge (ww), wall-ceiling edge (wc)\}.

\begin{figure}
\begin{center}
\includegraphics[width=8.5cm]{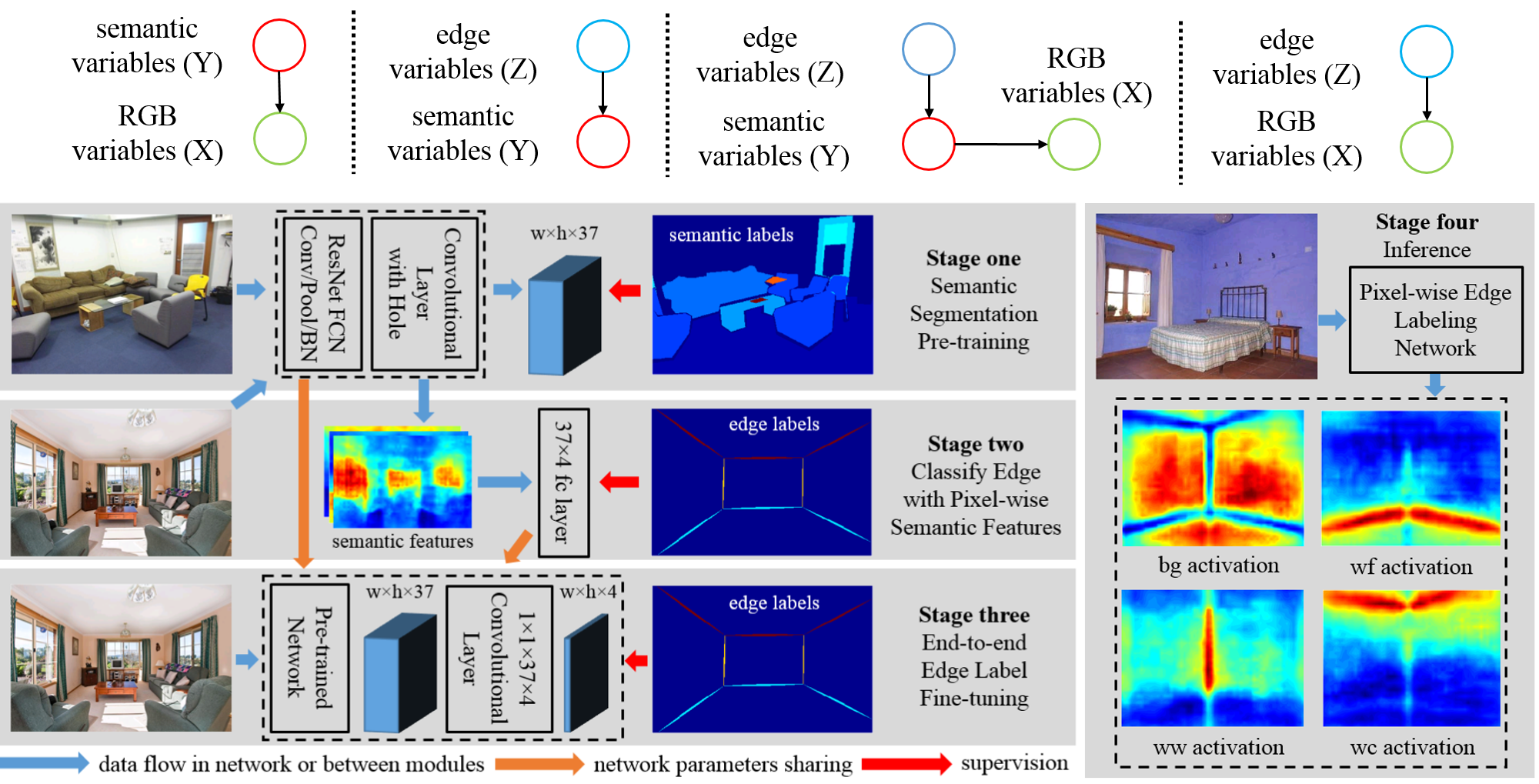}
\end{center}
   \caption{Top: Probabilistic node connectivity. Bottom: Semantic transfer. In stage three, \emph{pre-trained network} refers to the one outlined by the dashed box in stage one. In stage four, \emph{pixel-wise edge labelling network} refers to the one outlined by the dashed box in stage three. Better viewed electronically for a higher resolution.}
\label{fig:overview}
\end{figure}

\begin{equation}\label{es}
S = \{bg, wf, ww, wc\}
\end{equation}

 Activations of the last layer are incorporated into the conventional inference framework as features. \cite{dasgupta2016delay} uses another formulation in which every pixel may be assigned a label from a 5-class set \{floor, left wall, middle wall, right wall, ceiling\}. This 5-class formulation has an ambiguity problem as the patterns of three type of walls are not discriminative in nature. FCN is coordinate-invariant since convolutional layers actually conduct a sliding window search, so it is not suitable to tell the difference between \emph{left wall} and \emph{right wall}. Thus \cite{dasgupta2016delay} uses an additional ambiguity clarification step. \cite{ren2016cfile} uses both formulations for FCN training. These FCN-based works show dramatic performance improvements but as stated by the second motivation, their inference schemes remain conventional or primitive. With robust FCN features, it is possible to design more principled and efficient inference schemes.

\textbf{Broader literature.} There are actually other scene understanding tasks substantially same as or similar to room layout estimation. For example, \cite{nedovic2007depth} tries to understand the layouts of natural scenes with a horizon, urban scenes, corridors and others, for which room layout estimation is only a special case. Another special case of \cite{nedovic2007depth} is \emph{outdoor urban layout estimation}, such as \cite{barinova2008fast}\cite{hoiem2007recovering}. It is often regarded as a graphics application under the name of \emph{photo pop-up} and evaluated with subjective user study. \cite{lee2009geometric} tries to recover more detailed room layout than a box and evaluates with wall-floor edge error. Since these works exploit techniques that \cite{hedau2009recovering} is built upon, they could potentially benefit from the method proposed in this paper.

\textbf{Concepts similar to ST and PIO.} If we look at an even broader literature, the concepts somewhat similar to ST and PIO have already been discussed. Under the name of label transfer, \cite{liu2009nonparametric}\cite{zhang2010supervised} address semantic segmentation in a nonparametric manner. ST is different from them primarily as a unified deep architecture (and of course in its parametric nature). \cite{felzenszwalb2005pictorial} and its followers are famous for stating human limbs as springs. PIO is different from them primarily as an efficient approximation inspired by mechanics concepts.


\begin{figure*}
\begin{center}
\includegraphics[width=17cm]{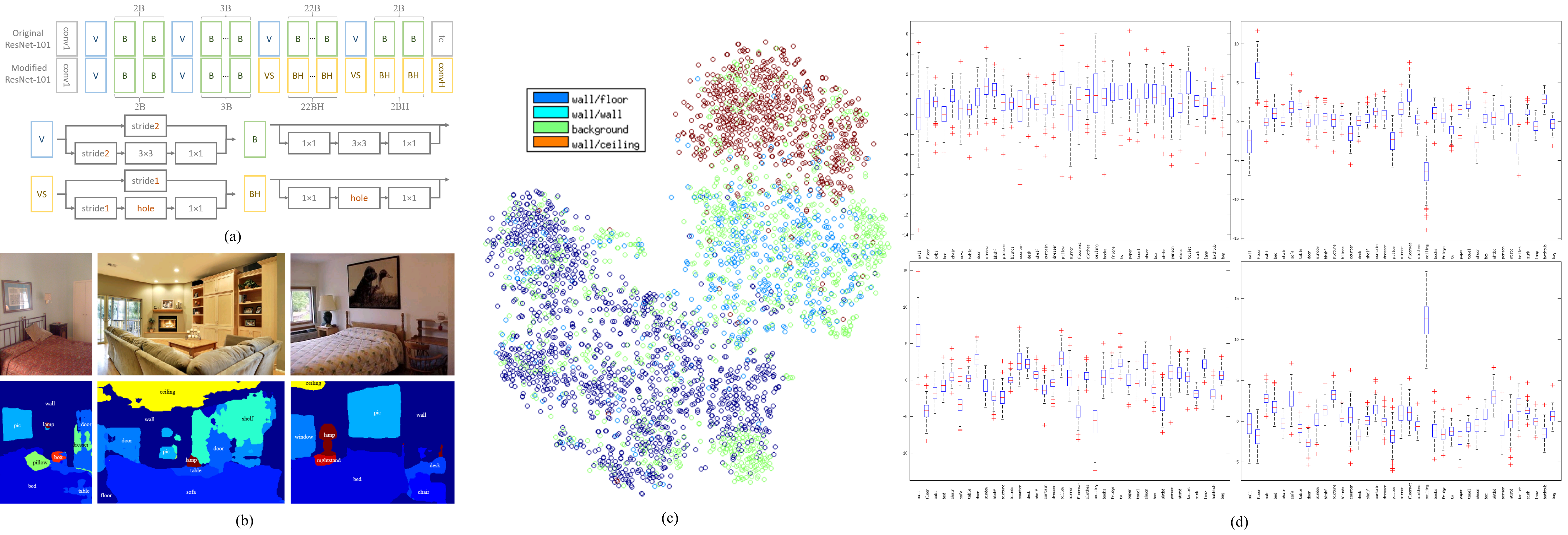}
\end{center}
   \caption{(a) Network design for ST stage one. (b) Qualitative results for semantic segmentation on dataset LSUN. Note that LSUN does not provide semantic segmentation ground truth. (c) Unsupervised structure visualization of the semantic feature space. (d) Transfer weights Visualization. Left-top: bg. Right-top: wf. Left-bottom: ww. Right-bottom: wc. Better viewed electronically for a higher resolution.}
\label{fig:stex}
\end{figure*}

\section{Semantic Transfer}

Here we present semantic transfer which is made up of 3 stages (Fig~\ref{fig:overview}). Firstly we look at the inference phase: the ultimate goal of our FCN is pixel-wise edge labelling. As demonstrated by Fig~\ref{fig:overview}'s \emph{stage four} panel, four pixel-wise activation maps are extracted from the input image, with each one corresponding to a label from $S$ (set~\ref{es}). For example, in the \emph{wf activation} map higher color temperature indicates higher possibility of \emph{wf} existence.

In stage one, we train an FCN for 37-class semantic segmentation on dataset SUNRGBD in order to describe a cluttered scene to the utmost extent. These 37 categories can cover most of the stuff and furniture that commonly appear in an indoor scene, like wall, ceiling, chair or window. We build this FCN upon the newly introduced architecture ResNet-101 \cite{he2016deep}. As Fig~\ref{fig:stex}a shows, we do net surgeries to the last two sets of bottlenecks in original ResNet-101, with the \emph{hole} mechanism described in \cite{chen14semantic} (under the name of dilated convolution in \cite{yu2015multi}). Inputs to this network (RGB images) are actually random variables $X$ taking values from $[0,255]$. $X$ is determined by hidden random variables $Y$ taking values from semantic labels $[1,37]$. Thus this network describes the posterior distribution $P(Y|X)$.


In stage two, we feed the room layout dataset LSUN through the semantic segmentation network, producing pixel-wise 37-channel semantic features. Since they are both indoor scene understanding datasets, the model trained on SUNRGBD generalizes well on LSUN. Fig~\ref{fig:stex}b shows some qualitative results on LSUN, all of which are produced by a softmax operation without post-processing techniques like conditional random fields. Then treating every pixel as a sample, we learn a fully connected layer to bridge the gap between 37-channel semantic features and 4-class edge labels. In order to illustrate that semantic features are discriminative for this task, we do a standard unsupervised analysis with t-sne \cite{van2008visualizing}. As Fig~\ref{fig:stex}c shows, samples of wall-ceiling edges (wc) and wall-floor edges (wf) form obvious clusters in the embedding space. Yet some samples of wall-wall edges (ww) and background (bg) scatter among each other. In this stage, $Y$ is determined by hidden random varibles $Z$ taking values from edge labels $[1,4]$ (set~\ref{es}). So this fc layer describes the posterior distribution $P(Z|Y)$.

$P(Z|Y)$ is a parameterized representation of the relationship between room layout and scene clutter. Unlike pioneering works, we model this relationship directly in a neural network. This is inspired by how a human understands room layout. As demonstrated by Fig~\ref{fig:overview}'s \emph{stage two} panel, the network in stage one extracts 37-channel semantic features from the scene. Only the channel on top of the stack is fully illustrated, and that channel corresponds to \emph{window}. This channel can roughly tell the locations and extensions of three windows in the scene. How would a human brain parse room layout from semantic features like this? We hypothesize that it makes decisions according to rules like:

\emph{wall-floor edges cannot go through windows, so they are less possible to appear in areas with high window scores}.

In order to validate that the network behaviors are consistent to this hypothesis, we visualize the transfer weights in this fc layer. These weights are learnt for 100 times independently and organized into box figure as Fig~\ref{fig:stex}d. Not surprisingly, \emph{wall}, \emph{floor} and \emph{ceiling} channels of semantic features contribute the most to \emph{ww}, \emph{wf} and \emph{wc}, respectively. Generally speaking, higher scores with smaller boxes means stronger correlation. We take \emph{wc} for example. Except for \emph{ceiling}, top four transfer weights come with \emph{cabinet}, \emph{picture}, \emph{sofa} and \emph{whiteboard}. According to common sense, \emph{cabinet}, \emph{picture} and \emph{whiteboard} tend to appear in the receptive field of a \emph{wc} pixel, because they are vertically high in physical space. \emph{sofa} is usually lower, so its variation (depicted by box size) is twice \emph{picture}'s. \emph{whiteboard} is rare, explaining why its variation is also large.

In stage three, this learnt $37\times4$ fc layer is reshaped into a $1\times1\times37\times4$ convolutional layer and added on top of the network trained in stage one. Weights from stage one act as a feature extractor, and weights from stage two act as a classifier. They form a pixel-wise edge labelling network describing $P(Z|Y)P(Y|X)=P(Z|X)$. On one hand, this network can be end-to-end fine-tuned on LSUN for edge labelling, which is the ultimate goal we mentioned at the beginning of this section. On the other hand, it incorporates the relationship between scene clutter and room layout elegantly, which is the first motivation of this paper.

Except for end-to-end training and scene clutter modelling, another advantage of semantic transfer is better initialization for extremely unbalanced training data. We have tried to train this pixel-wise edge labelling network directly with the ResNet FCN (Fig~\ref{fig:stex}a), leaving out ST. But outputs of the batch normalization (BN) layers are prone to overflow, making training fail. Training problems are also reported in \cite{mallya2015learning}'s 3.2 section. It says the network has to be pre-trained on NYUd2 and pre-training on PASCAL leads to bad results. This problem may be caused by the extremely unbalanced distribution of edge labels. As shown by Fig~\ref{fig:overview}'s \emph{stage two} panel, over 99\% labels are \emph{background}. Like the classical method of initializing an auto-encoder with multiple restricted boltzmann machines \cite{hinton2006reducing}, our pixel-wise edge labelling network is initialized by the first two stages. We no longer observe the overflow phenomenon with ST.

Probabilistic nodes' connectivity is illustrated by Fig~\ref{fig:overview}'s upper part. The pre-trained model will be released. Details about network, unsupervised analysis and weights visualization are provided in the supplementary material.

%

\section{Optimization}

We provide comprehensive feature quality visualizations and comparisons in the supplementary material. For parameterized room layout inference, we propose two techniques: naive optimization (NO) and its efficient approximation named physics inspired optimization (PIO).

\begin{figure}
\begin{center}
\includegraphics[width=8.5cm]{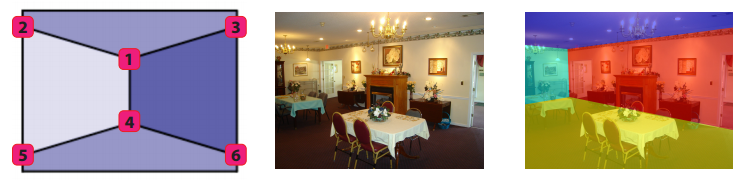}
\end{center}
   \caption{The 6th topology, clipped from LSUN specification.}
\label{fig:top}
\end{figure}

There are 11 different possible room layout topologies in a 2D image, as demonstrated in the supplementary material. We index them with $i$. Each topology is parameterized by the edge conjunctions set $P_i=\{P_{ij},j\in[1,nC]\}$, with every $P_{ij}$ as a 2D coordinate and $nC$ as the conjunction number. Conjunction connectivity is defined by edge set $E_i=\{E_{ik}=(Q_{ka},Q_{kb},c),Q_{ka}\in P_i,Q_{kb}\in P_i,c\in\emph{S},k\in[1,nE]\}$, with $nE$ as the edge number. $S$ is set~\ref{es}. The 6th topology is demonstrated in Fig~\ref{fig:top} as an example. $P_{i}$ and $E_{i}$ can be converted into a pixel-wise edge label map which is similar to the output in Fig~\ref{fig:problem}. This \textbf{conversion} is denoted as $M=C(P_{i},E_{i})$ and we will omit $E_{i}$ later because it does not change for a certain topology. Also, we will use $M[P_{i}]$ when referring the map $M$ produced from conjunction set $P_i$ and $M[P_{ij}]$ when a certain conjunction $P_{ij}$ is under consideration.

The features produced by the pixel-wise edge labelling network are denoted as $F_l(l\in[1,4])$. Note that both $M$ and $F_l$ are of the same size as input image, denoted by $(w,h)$. On them we define the consistency objective ($CO$) and its corresponding energy format ($e$):

\begin{equation}\label{co}
CO = \frac{1}{wh}\sum_{l=1}^{4}\sum_{m=1}^{w}\sum_{n=1}^{h}F_l(m,n)\times M_l(m,n)
\end{equation}

\begin{equation}
e = \exp(-CO)
\end{equation}
in which $M_l(l\in[1,4])$ is the binary mask generated from $M$ by setting a pixel to one if $M(m,n)=l$ and zero otherwise. For every different topology we can find the best parameterized representation $P_{i}$ by minimizing $e$:

\begin{equation}\label{obj}
\overline{P_{i}} = \arg\min_{P_{i}} e
\end{equation}

In most cases, starting from the right topology leads to the lowest energy value and wrong topologies lead to higher energy values. Failure cases do exist and we will visualize them later.
All optimization implementations detailed below are initialized from the average state of $P_{i}$ set (such as the one demonstrated by Fig~\ref{fig:top}'s left figure).

\subsection{Naive Optimization}

To solve Equation~\ref{obj}, firstly we propose NO as follows:

\begin{equation}\label{npo1}
\frac{\partial e}{\partial P_{ijx}}\approx e(P_{ij(x+\Delta x)})-e(P_{ij(x-\Delta x)})
\end{equation}

\begin{equation}\label{npo2}
\frac{\partial e}{\partial P_{ijy}}\approx e(P_{ij(y+\Delta y)})-e(P_{ij(y-\Delta y)})
\end{equation}

\begin{equation}\label{npo3}
\Delta P_{ij} = \alpha\times(-\frac{\partial e}{\partial P_{ijx}},-\frac{\partial e}{\partial P_{ijy}})
\end{equation}

\begin{algorithm}
\caption{Naive Optimization}
\begin{algorithmic}
\STATE \textbf{Initialize:} average $P_{i}$
\WHILE{$e$ decreases}
\FORALL{$j$}
\STATE update $P_{ij}$ according to Equation~\ref{npo1}, ~\ref{npo2}, and ~\ref{npo3}
\ENDFOR
\STATE calculate $e$ at updated $P_{i}$
\ENDWHILE
\end{algorithmic}
\label{npoalgo}
\end{algorithm}
in which $\alpha$ is the scaling factor and $\Delta x$ ($=\Delta y$) is the window size. For conjunctions at image boundary (e.g. $P_{62}$ in Fig~\ref{fig:top}'s left figure), an additional constraint is imposed by setting corresponding component of $\Delta P_{ij}$ to zero. If conjunctions move to image corners, $\Delta P_{ij}$ is treated as a special case so as to allow the conjunction to move onto another boundary or just stick to the corner. The convergence performance of NO is good but it is very slow so we introduce PIO as an efficient alternative.

\subsection{Analysis and Motivation}

We first analyze the efficiency bottleneck of NO. When calculating Equation~\ref{npo1} (and similarly Equation~\ref{npo2}),

\begin{equation}\label{pio1}
\frac{\partial e}{\partial P_{ijx}}\propto-(CO(P_{ij(x+\Delta x)})-CO(P_{ij(x-\Delta x)}))
\end{equation}
\begin{equation}\label{pio2}
=-\sum_{l=1}^{4}\sum_{m,n}F_l\times (M_l[P_{ij(x+\Delta x)}]-M_l[P_{ij(x-\Delta x)}])
\end{equation}

\begin{figure}
\begin{center}
\includegraphics[width=8.5cm]{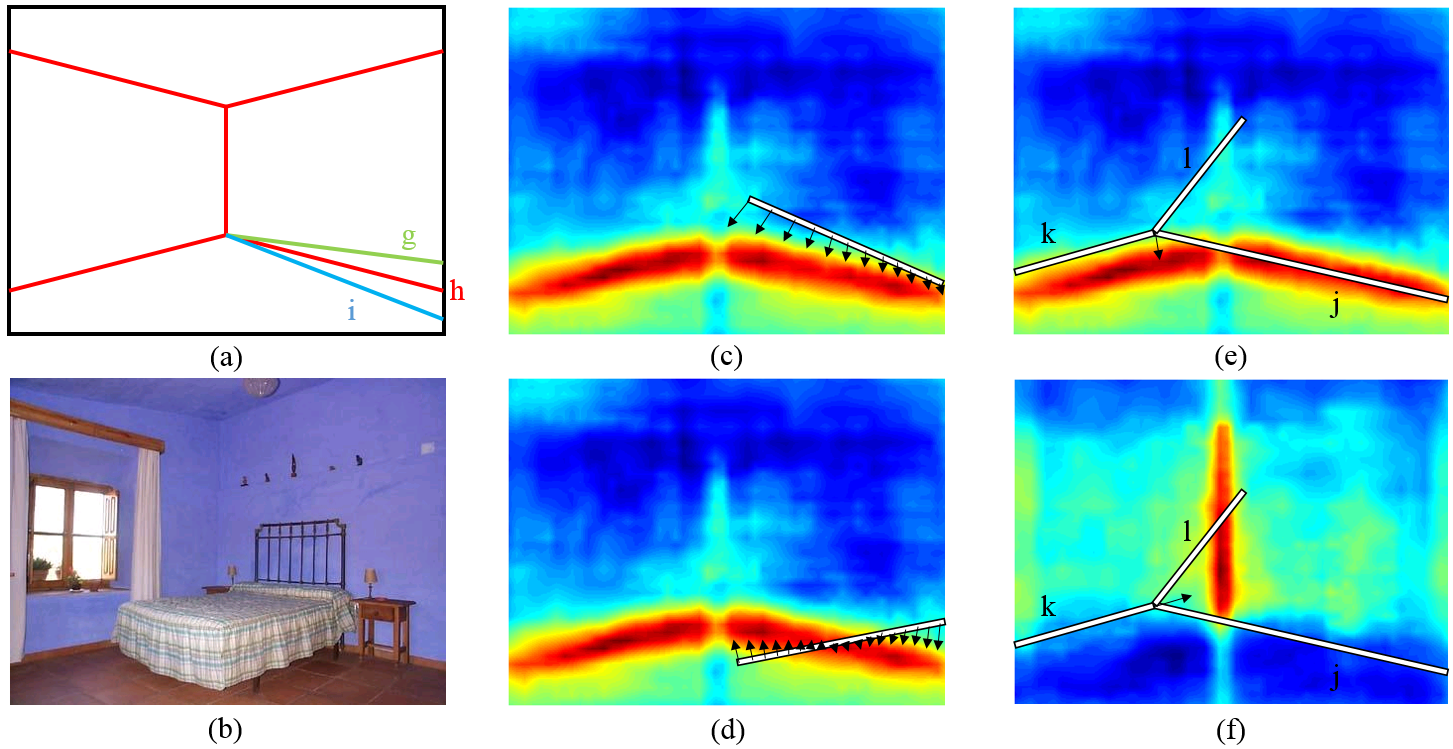}
\end{center}
   \caption{(ag) $M_2[P_{ij(x-\Delta x)}]$. (ah) $M_2[P_{ij}]$. (ai) $M_2[P_{ij(x+\Delta x)}]$. (b) Input image. (c/d) If we consider an edge as a spring and the feature map as a potential field, forces imposed on the spring's every point are correlated. (e/f) The influence of force composition.}
\label{fig:physics}
\end{figure}

In Equation~\ref{pio2}, we omit $m,n$ whose meanings are stated in Equation ~\ref{co}. Calculating $M_l[P_{ij(x+\Delta x)}]-M_l[P_{ij(x-\Delta x)}]$ is the efficiency bottleneck, which represents $M=C(P_{i})$'s gradient and we illustrate $M_2[P_{ij(x+\Delta x)}]-M_2[P_{ij(x-\Delta x)}]$ with Fig~\ref{fig:physics}a, which subtracts two pixel-wise masks.

As a reminder, $M$ is generated by conversion $C$. At first, we implement $C$ by traversing every pixel to decide its label. If we denote the scale of $w,h$ by N, the complexity of this implementation (referred as NOA later) is $O(N^2)$ for every calculation of Equation~\ref{npo1} or~\ref{npo2}. It runs for tens of minutes for an image. An improved implementation (referred as NOB later) of $C$ calculates pixel coordinates between two conjunctions and accesses corresponding mask element directly. Its complexity is $O(N)$, and it runs for about 30s for an image. The idea of further reducing the complexity to $O(1)$ motivates us to introduce PIO.

We consider every edge as a spring which may translate, rotate and change its length. In NO, edges' movements are decided by every pixel on them, yet there is computation redundance. As demonstrated by Fig~\ref{fig:physics}c and Fig~\ref{fig:physics}d, we consider the feature map as a potential field and analyze how points on the edge move. Not surprisingly, their movements are not independent and can be roughly interpolated from the movements of the edge's two endpoints, that is $Q_{ka}$ and $Q_{kb}$. Based on this observation, we propose to approximate $\Delta P_{ij}$ with gradients defined on $P_{i}$ instead of $M_l[P_{i}]$. Since the number $nC$ of conjunctions $P_{i}$ is constant, the complexity is $O(1)$. This is PIO's first key concept.

Force composition is the second key concept of PIO. As demonstrated by Fig~\ref{fig:physics}e, if we consider the endpoints of edge j and k instead of every points on them they will move towards a local minima state. This will be corrected by calculating the movements of edge l (Fig~\ref{fig:physics}f), in which another feature map (wall-wall edge) will be used as the potential field. So the movement of every conjunction should be decided by the forces imposed on every edge that is connected to that conjunction. Obviously adding two gradient vectors (such as the ones in Fig~\ref{fig:physics}e and Fig~\ref{fig:physics}f) naturally obeys the parallelogram law of force composition.

\begin{figure*}
\begin{center}
\includegraphics[width=17cm]{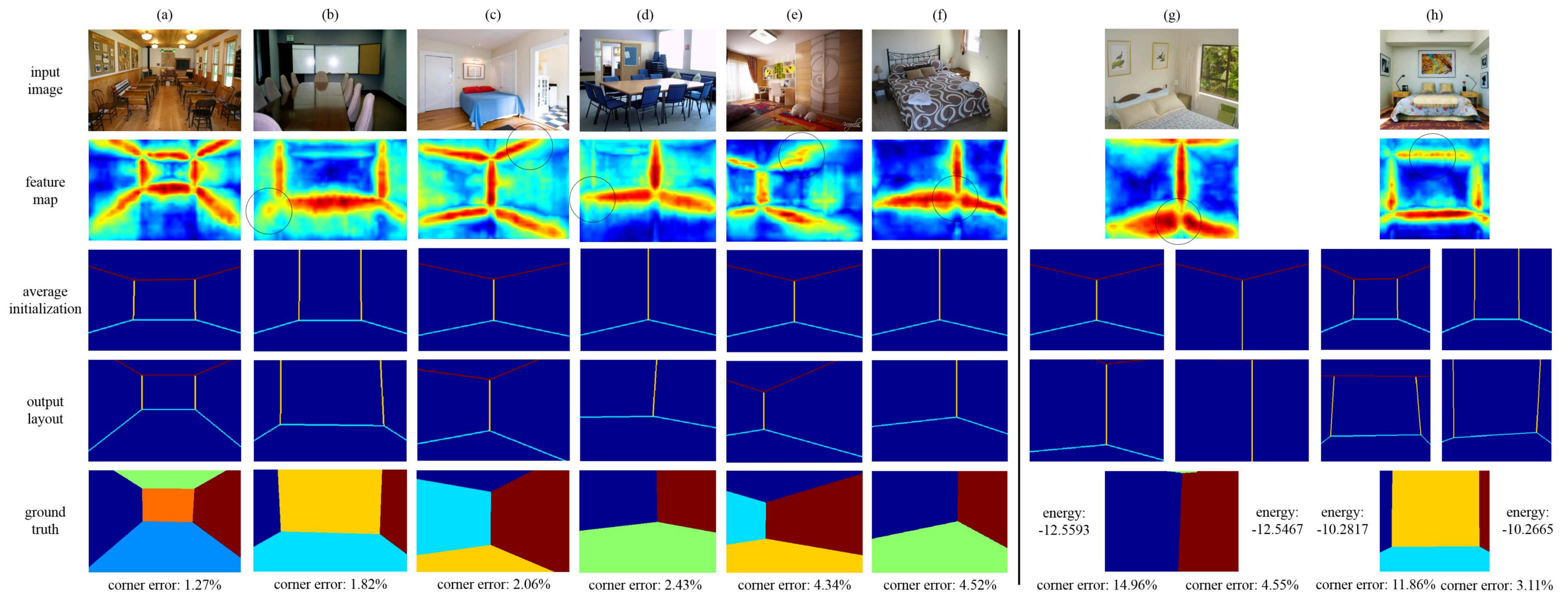}
\end{center}
   \caption{Left: qualitative results on LSUN validation set. The visualized feature map merges wf, ww, and wc by a pixel-wise max operation, yet they are used independently in PIO. Right: typical failure cases in which a wrong topology produces the lowest energy.}
\label{fig:finalquali}
\end{figure*}

\subsection{Physics Inspired Optimization}

For the first concept, we define a new consistency objective for each endpoint of an edge $E_{ik}=(Q_{ka},Q_{kb},c)$:

\begin{equation}\label{co2}
CO2 = F_c(Q_{kax},Q_{kay})
\end{equation}

\begin{equation}
e2 = \exp(-CO2)
\end{equation}

As a reminder, the meanings of $E$ and $F$ are stated at the beginning of this section. Calculating the gradient of a certain point in a potential field is trivial as:

\begin{equation}\label{pio3}
\frac{\partial e2}{\partial Q_{kax}}\approx e2(Q_{ka(x+\Delta x)})-e2(Q_{ka(x-\Delta x)})
\end{equation}

\begin{equation}\label{pio4}
\frac{\partial e2}{\partial Q_{kay}}\approx e2(Q_{ka(y+\Delta y)})-e2(Q_{ka(y-\Delta y)})
\end{equation}

\begin{equation}\label{pio5}
\Delta Q_{ka} = \alpha\times(-\frac{\partial e2}{\partial Q_{kax}},-\frac{\partial e2}{\partial Q_{kay}})
\end{equation}

In this physics inspired optimization, $\Delta Q_{ka}$ is regarded as a force imposed on the endpoint $Q_{ka}$ of an spring-like edge $E_{ik}$. As for the second concept of force composition, we define $E[P_{ij}]=\{(Q_{oa}=P_{ij},Q_{ob},c),o\in[1,\#(E[P_{ij}])]\}$ which is a subset of $E_{i}$.
And the force imposed on $P_{ij}$ when considering different edges can be denoted as $\Delta Q_{oa}$. Thus we approximate $\Delta P_{ij}$ with:

\begin{equation}\label{pio6}
\Delta P_{ij} = \sum_{o=1}^{\#(E[P_{ij}])}\Delta Q_{oa}
\end{equation}

\begin{algorithm}
\caption{Physics Inspired Optimization}
\begin{algorithmic}
\STATE \textbf{Initialize:} average $P_{i}$
\WHILE{$e$ decreases}
\FORALL{$j$}
\STATE get the subset $E[P_{ij}$]
\FORALL{$o$}
\STATE calculate the force imposed on $Q_{oa}=P_{ij}$ according to Equation~\ref{pio3}~\ref{pio4}~\ref{pio5}
\ENDFOR
\STATE calculate $\Delta P_{ij}$ by Equation~\ref{pio6}, and update $P_{ij}$
\ENDFOR
\STATE calculate $e$ at updated $P_{i}$
\ENDWHILE
\end{algorithmic}
\label{pioalgo}
\end{algorithm}

As mentioned before, Equation~\ref{pio6} naturally obeys the parallelogram law of force composition. In case of potential confusion, we clarify that both $\Delta Q_{oa}$ and $\Delta Q_{ka}$ are calculated according to Equation ~\ref{pio5} ($k$ is the index in $E_{i}$ and $o$ is the index in $E_{i}$'s subset $E[P_{ij}]$). And to summarize, PIO's efficiency primarily comes from Equation~\ref{co2}'s $O(1)$ complexity while Equation~\ref{co}'s is at least $O(N)$ (NOB).

\section{Experiments}

\subsection{LSUN Results}

LSUN is a room layout estimation dataset consisted of 4000 training, 394 validation, and 1000 held-out testing samples. Two standard metrics are used for evaluation: (1) $\textbf{e}_{\textbf{corner}}$. Corner (conjunction) error is the Euclidean distance between estimated coordinates of $P_i$ and ground truth. Because of resolution diversity, $\textbf{e}_{\textbf{corner}}$ is normalized by image diagonal length. (2) $\textbf{e}_{\textbf{pixel}}$. By converting $P_i$ into mask representation like the ground truth in Fig~\ref{fig:finalquali}, pixel error measures the ratio of mislabelled pixels to all pixels. (For $\textbf{e}_{\textbf{pixel}}$'s label ambiguity problem, LSUN official evaluation codes automatically maximize the overlap.)

For a large-scale evaluation, both metrics are averaged over images. On validation set, official evaluation codes provided by LSUN committee are used. Third-party evaluation results on the test set are reported in Table~\ref{lsuntest}. The proposed method outperforms conventional method \cite{hedau2009recovering} and FCN-based methods \cite{mallya2015learning}\cite{dasgupta2016delay}\cite{ren2016cfile} on both metrics. Qualitative results and failure cases on validation set are demonstrated by Fig~\ref{fig:finalquali}. Eight videos showing how PIO works are provided in the supplementary material, with each one corresponding a sample in Fig~\ref{fig:finalquali}, respectively.

Fig~\ref{fig:finalquali}a shows a typical easy case, in which most edge pixels are visible and the feature map captures their locations accurately. As \emph{video-a.wmv} shows, the visualized edge map gets twisted temporarily near 30th iteration because of force composition and PIO finally aligns it with the true layout.

Fig~\ref{fig:finalquali}bcd show some cases in which the feature maps fail to locate edges accurately where the black circles are, leading to relatively higher $\textbf{e}_{\textbf{corner}}$. Reasons are diverse, such as severe occlusion (b), insufficient feature map resolution (c), and misleading texture (d).

In Fig~\ref{fig:finalquali}e, the room is no longer a strict box if we consider the cabinet as a part of wall. Actually those separated wall-ceiling edges are successfully captured by the feature map and aligned by PIO. However, the annotation protocol takes the cabinet as occlusion.  Fig~\ref{fig:finalquali}f shows a heavily-occluded case. Semantic transfer allows the network to extrapolate the existence of wall-floor edges behind the bed, but the conjunction in the black circle is not accurately localized.

Although not 100\% accurate, Fig~\ref{fig:finalquali}a-f are regarded as successful cases as the output topologies are right. Fig~\ref{fig:finalquali}gh are two typical failure cases in which a wrong topology produces the lowest energy. Fig~\ref{fig:finalquali}g's failure is caused by over-fitting. The network extrapolates there are wall-floor edges behind the bed, yet the annotation protocol does not. \emph{video-g1.wmv} shows the optimization procedure of the wrong topology and \emph{video-g2.wmv} shows that of the right topology. Even though the latter leads to a lower error but the algorithm outputs the former as it produces a lower energy. Fig~\ref{fig:finalquali}h demonstrates another type of failures caused by structure ambiguity. Again this scene is no longer a strict box as some parts of the wall protrude outwards. The network recognizes them as ceiling but the annotation protocol does not, causing PIO to output a wrong topology.

\subsection{Hedau Results}

\begin{table}[t]
  \centering
  \centering
  \begin{tabular}{p{3.5cm}p{1.5cm}p{1.5cm}}
    \textbf{Method} & $\textbf{e}_{\textbf{pixel}}$ (\%) & $\textbf{e}_{\textbf{corner}}$ (\%)\\[0.05cm]
    \hline\\[-0.2cm]
    Hedau et al.(2009) \cite{hedau2009recovering}  & 24.23 & 15.48\\[0.05cm]
    Mallya et al.(2015) \cite{mallya2015learning}  & 16.71 & 11.02\\[0.05cm]
    Dasgupta et al.(2016) \cite{dasgupta2016delay}  & 10.63 & 8.20\\[0.05cm]
    Ren et al.(2016) \cite{ren2016cfile} & 7.57 & 5.23\\[0.05cm]
    Ours & \textbf{5.29} & \textbf{3.84}\\[0.05cm]
    \hline\\[-0.2cm]
    \end{tabular}

    \caption{Quantitative results on LSUN test set.}
    \label{lsuntest}
\end{table}

\begin{figure}
\begin{center}
\includegraphics[width=8cm]{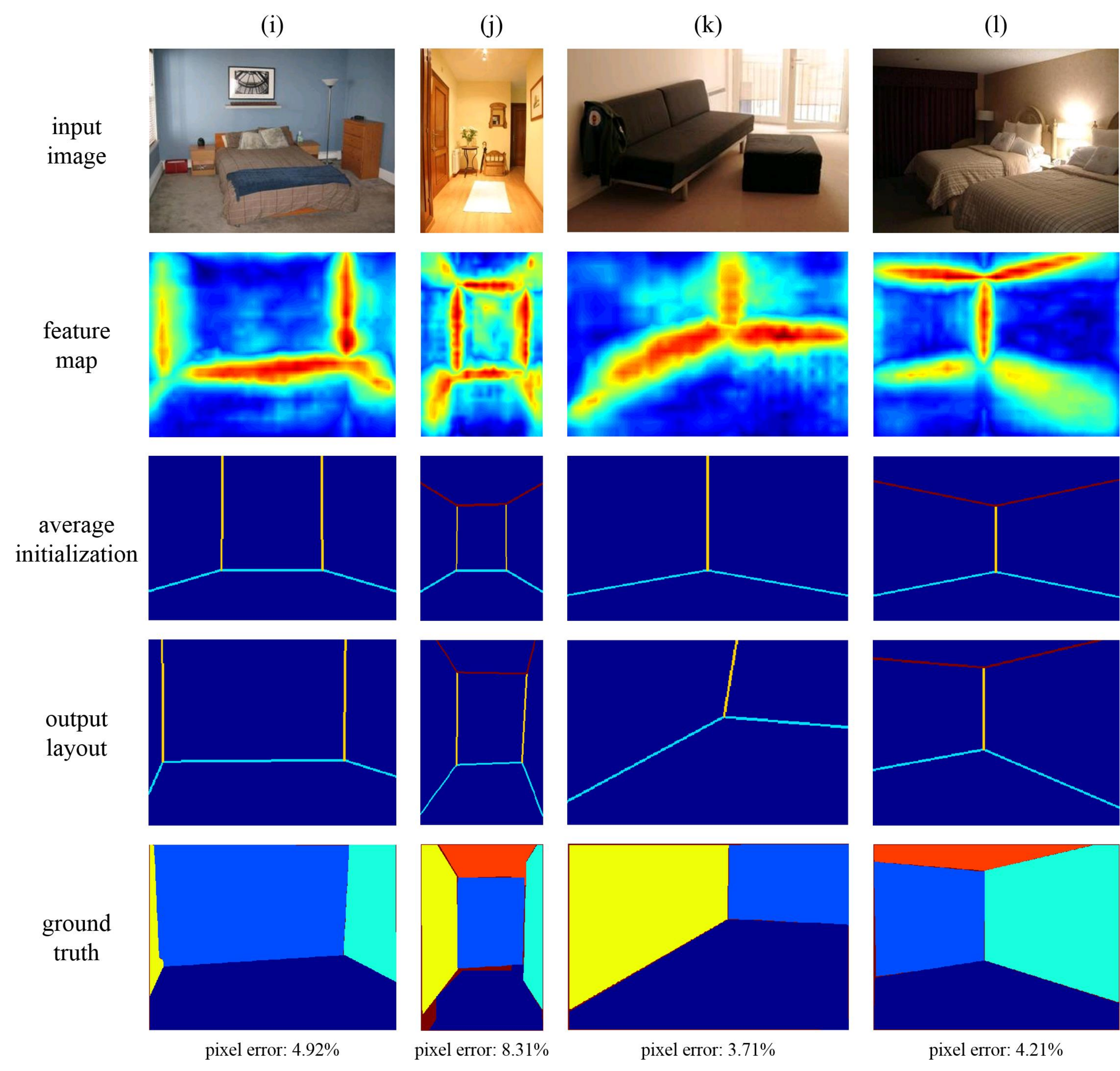}
\end{center}
   \caption{Qualitative results on Hedau test set.}
\label{fig:hedauquali}
\end{figure}

\begin{table}[t]
  \centering
  \centering
  \begin{tabular}{p{4.5cm}p{1.5cm}}
    \textbf{Method} & $\textbf{e}_{\textbf{pixel}}$ (\%)\\[0.05cm]
    \hline\\[-0.2cm]
    Hedau et al.(2009) \cite{hedau2009recovering}  & 21.20\\[0.05cm]
    Del Pero et al.(2013) \cite{del2013understanding}  & 12.70\\[0.05cm]
    Mallya et al.(2015) \cite{mallya2015learning}  & 12.83\\[0.05cm]
    Dasgupta et al.(2016) \cite{dasgupta2016delay}  & 9.73\\[0.05cm]
    Ren et al.(2016) \cite{ren2016cfile} & 8.67\\[0.05cm]
    Ours & \textbf{6.60}\\[0.05cm]
    \hline\\[-0.2cm]
    \end{tabular}

    \caption{Quantitative results on Hedau test set. To clarify, \cite{hedau2009recovering}\cite{del2013understanding} are not trained on the large-scale dataset LSUN.}
    \label{hedautest}
\end{table}

The Hedau dataset is presented by \cite{hedau2009recovering}, being consisted of 209 training samples and 105 testing samples. On Hedau test set, We directly evaluate the model trained with LSUN training set. As Fig~\ref{fig:hedauquali} shows, this model extracts reliable features across datasets. Consistent to the literature we use pixel error as quantitative metric. We report better results than conventional methods like \cite{hedau2009recovering}\cite{del2013understanding} and FCN-based methods like \cite{mallya2015learning}\cite{dasgupta2016delay}\cite{ren2016cfile} (Table ~\ref{hedautest}). Overall pixel error (6.60\%) on Hedau test set is higher than that (5.29\%) on LSUN test set because the ground truth mask annotated by Hedau dataset is more strict (typically shown by Fig~\ref{fig:hedauquali}j).

\subsection{Hyper Parameters and Efficiency}

\begin{figure}
\begin{center}
\includegraphics[width=7cm]{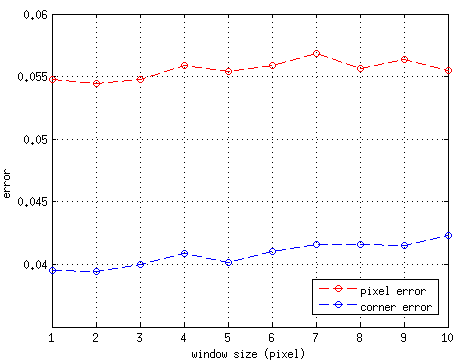}
\end{center}
   \caption{LSUN validation set error v.s. window size.}
\label{fig:window}
\end{figure}

(1) In Algorithm~\ref{pioalgo}, wether $e$ decreases is determined by a threshold of $10^{-6}$. This threshold is related to the numerical scale of $e$. During implementation, we use $e=-CO$ instead of $e=\exp(-CO)$ because of equivalence, and $e$'s numerical scale is around $-15$ which has already been shown in videos mentioned above. (2) Scaling factor $\alpha$ is self-adaptive to ensure that the gradients' (forces') length is between 1 and 3. This restricts the conjunction to move only a little in one iteration, as the videos show. (3) The influence of window size $\Delta x$ ($=\Delta y$) is evaluated on LSUN validation set, and the quantitative results are demonstrated by Fig~\ref{fig:window}. As the window size grows from 1 pixel to 10 pixels, both metrics show a trend of increase. Since PIO can be regarded as an alignment algorithm, this is not surprising because calculating gradients in a larger window size leads to a weaker ability to accurately capture local structure.

\begin{table}[t]
  \centering
  \centering
  \begin{tabular}{p{1.5cm}p{0.8cm}p{0.8cm}p{0.8cm}}
     & NOB & PIO\\[0.05cm]
    \hline\\[-0.2cm]
    \textbf{artpf} (s) & 35.41 & 1.79\\[0.05cm]
    $\textbf{e}_{\textbf{pixel}}$ (\%) & 5.42 & 5.48\\[0.05cm]
    $\textbf{e}_{\textbf{corner}}$ (\%) & 3.88 & 3.95\\[0.05cm]
    \hline\\[-0.2cm]
    \end{tabular}

    \caption{Average running time per frame (artpf) comparison.}
    \label{time}
\end{table}

We evaluate average running time per frame on LSUN validation set with NOB and PIO. NOA is not evaluated because of intractability. The results are provided in Table~\ref{time}, showing that PIO brings dramatic speedup against NOB without causing noticeable accuracy loss. Our codes are all implemented with MATLAB, thus there is still much head room for potential real-time applications.

\subsection{Ablative Study of ST}

In order to evaluate the impact of semantic transfer, we use three standard edge prediction accuracy metrics as follows: F-score @ optimal dataset scale (ODS), F-score @ optimal image scale (OIS), average precision (AP). We consider following settings: (A) directly train a VGG16-based network for edge labelling. (B) train a VGG16-based network with semantic transfer. (C) train a Resnet101-based network with semantic transfer. All settings (including D-G in next subsection) use same hyperparameters. As shown by Table ~\ref{edge}, setting B comes with a higher accuracy than A ($2.8\%  \Delta$ODS), and setting C sees a larger improvement ($4.3\%  \Delta$ODS). This indicates that both semantic transfer and the introduction of Resnet101 bring improvements yet the latter takes a relatively larger part.

\subsection{Representation Learning Perspective}

In order to further compare semantic transfer with traditional representation learning schemes, we consider these settings (all on a VGG16-based network): (D) pre-train on SUNRGBD for semantic segmentation, re-initialize the last layer and fine-tune all parameters. (E) same as D except that we only fine-tune parameters after layer 5b. (F) semantic transfer and in stage three fine-tune all parameters (It is same as B). (G) same as F except that we only fine-tune parameters after layer 5b. As shown by Table ~\ref{edge}, F's performance is slightly better than D ($0.3\%  \Delta$ODS, one may argue this could be caused by additional parameters or stochastic training) yet this margin gets more significant when we freeze parameters before layer 5b ($3.1\%  \Delta$ODS comparing G against E). This indicates that tuning a (properly initialized) $37\times4$ transfer layer is easier than re-training a (gaussian initialized) classification layer (more obvious in the frozen representation settings).

\begin{table}[t]
  \centering
  \centering
  \begin{tabular}{p{0.6cm}p{0.6cm}p{0.6cm}p{0.6cm}p{0.6cm}p{0.6cm}p{0.6cm}}
    & A & B(F) & C & D & E & G \\[0.05cm]
    \hline\\[-0.2cm]
    ODS & 0.243 & 0.271 & \textbf{0.314} & 0.268 & 0.202 & 0.233\\[0.05cm]
    OIS & 0.251 & 0.285 & \textbf{0.328} & 0.280 & 0.208 & 0.236\\[0.05cm]
    AP & 0.135 & 0.151 & \textbf{0.184} & 0.148 & 0.091 & 0.098\\ [0.05cm]
    \hline\\[-0.2cm]
    \end{tabular}

    \caption{Ablative study of ST on LSUN validation set.}
    \label{edge}
\end{table}

\begin{table}[t]
  \centering
  \centering
  \begin{tabular}{p{0.6cm}p{0.6cm}p{0.6cm}p{0.6cm}p{0.6cm}}
    & H & I & J & K \\[0.05cm]
    \hline\\[-0.15cm]
    $\textbf{e}_{\textbf{pixel}}$ & 11.28 &　6.31 & 5.75 & \textbf{5.48}\\[0.05cm]
    $\textbf{e}_{\textbf{corner}}$ & 8.55 & 4.98 & 4.17 & \textbf{3.95}\\[0.05cm]
    \hline\\[-0.15cm]
    \end{tabular}

    \caption{Ablative study of PIO on LSUN validation set.}
    \label{pio}
\end{table}

\subsection{Ablative Study of PIO}

With the classical pipeline (edge detection, vanishing point voting, ray sampling), we extract on average 334 proposals per image on LSUN validation set. Then with semantic transfer features (setting C above), we consider these settings: (H) pick the proposal that correlates to the features the most. (I) do PIO with the best proposal. (J) do PIO with top 10 best proposals and pick the one with the lowest energy. (K) the PIO setting mentioned above (without depending on those error-prone proposals generated from low-level edge cues). I sees a higher accuracy than H (-$4.97\% \Delta \textbf{e}_{\textbf{pixel}}$), which is not surprising as PIO refines layout proposals. Yet since I is restricted by the the proposal quality (which degenerates heavily in highly-occluded cases) thus K outperforms I (-$0.83\% \Delta \textbf{e}_{\textbf{pixel}}$). Augmenting with 10 proposals (J) sees a comparable performance with K. And generally speaking, PIO is better than ranking proposals (-$5.80\% \Delta \textbf{e}_{\textbf{pixel}}$ comparing K and H).

\section{Conclusion}

In this paper, we propose an alternative method for room layout estimation. With a very deep semantic transfer FCN, we extract reliable edge features under various circumstances. Meanwhile we develop PIO as a new inference scheme, which is inspired by mechanics concepts. The method's effectiveness is illustrated by extensive quantitative experiments on public datasets. Figures and videos are also provided as intuitive demonstrations.

\textbf{Acknowledgements.} This work was jointly supported by National Natural Science Foundation of China (Grant No.61132007, 61172125, 61601021, and U1533132).

{\small
\bibliographystyle{ieee}
\bibliography{egbib}

\begin{thebibliography}{10}\itemsep=-1pt

\bibitem{barinova2008fast}
O.~Barinova, V.~Konushin, A.~Yakubenko, K.~Lee, H.~Lim, and A.~Konushin.
\newblock Fast automatic single-view 3-d reconstruction of urban scenes.
\newblock In {\em ECCV 2008}.

\bibitem{coughlan1999manhattan}
J.~M. Coughlan and A.~L. Yuille.
\newblock Manhattan world: Compass direction from a single image by bayesian
  inference.
\newblock In {\em ICCV 1999}.

\bibitem{dasgupta2016delay}
S.~Dasgupta, K.~Fang, K.~Chen, and S.~Savarese.
\newblock Delay: Robust spatial layout estimation for cluttered indoor scenes.
\newblock In {\em CVPR 2016}.

\bibitem{del2012bayesian}
L.~Del~Pero, J.~Bowdish, D.~Fried, B.~Kermgard, E.~Hartley, and K.~Barnard.
\newblock Bayesian geometric modeling of indoor scenes.
\newblock In {\em CVPR 2012}.

\bibitem{del2013understanding}
L.~Del~Pero, J.~Bowdish, B.~Kermgard, E.~Hartley, and K.~Barnard.
\newblock Understanding bayesian rooms using composite 3d object models.
\newblock In {\em CVPR 2013}.

\bibitem{eigen2015predicting}
D.~Eigen and R.~Fergus.
\newblock Predicting depth, surface normals and semantic labels with a common
  multi-scale convolutional architecture.
\newblock In {\em CVPR 2015}.

\bibitem{eigen2014depth}
D.~Eigen, C.~Puhrsch, and R.~Fergus.
\newblock Depth map prediction from a single image using a multi-scale deep
  network.
\newblock In {\em NIPS 2014}.

\bibitem{felzenszwalb2005pictorial}
P.~F. Felzenszwalb and D.~P. Huttenlocher.
\newblock Pictorial structures for object recognition.
\newblock {\em IJCV 2005}.

\bibitem{gupta2015aligning}
S.~Gupta, P.~Arbel{\'a}ez, R.~Girshick, and J.~Malik.
\newblock Aligning 3d models to rgb-d images of cluttered scenes.
\newblock In {\em CVPR 2015}.

\bibitem{he2016deep}
K.~He, X.~Zhang, S.~Ren, and J.~Sun.
\newblock Deep residual learning for image recognition.
\newblock {\em CVPR 2016}.

\bibitem{hedau2009recovering}
V.~Hedau, D.~Hoiem, and D.~Forsyth.
\newblock Recovering the spatial layout of cluttered rooms.
\newblock In {\em CVPR 2009}.

\bibitem{hinton2006reducing}
G.~E. Hinton and R.~R. Salakhutdinov.
\newblock Reducing the dimensionality of data with neural networks.
\newblock In {\em Science 2006}.

\bibitem{hoiem2007recovering}
D.~Hoiem, A.~A. Efros, and M.~Hebert.
\newblock Recovering surface layout from an image.
\newblock {\em IJCV 2007}.

\bibitem{lee2009geometric}
D.~C. Lee, M.~Hebert, and T.~Kanade.
\newblock Geometric reasoning for single image structure recovery.
\newblock In {\em CVPR 2009}.

\bibitem{chen14semantic}
C.~Liang-Chieh, P.~George, K.~Iasonas, M.~Kevin, and L.~Y. Alan.
\newblock Semantic image segmentation with deep convolutional nets and fully
  connected crfs.
\newblock In {\em ICLR 2015}.

\bibitem{liu2009nonparametric}
C.~Liu, J.~Yuen, and A.~Torralba.
\newblock Nonparametric scene parsing: Label transfer via dense scene
  alignment.
\newblock In {\em CVPR 2009}.

\bibitem{mallya2015learning}
A.~Mallya and S.~Lazebnik.
\newblock Learning informative\ edge maps for indoor scene layout prediction.
\newblock In {\em ICCV 2015}.

\bibitem{nedovic2007depth}
V.~Nedovic, A.~W. Smeulders, A.~Redert, and J.-M. Geusebroek.
\newblock Depth information by stage classification.
\newblock In {\em ICCV 2007}.

\bibitem{ramalingam2013manhattan}
S.~Ramalingam, J.~K. Pillai, A.~Jain, and Y.~Taguchi.
\newblock Manhattan junction catalogue for spatial reasoning of indoor scenes.
\newblock In {\em CVPR 2013}.

\bibitem{schwing2013box}
A.~G. Schwing, S.~Fidler, M.~Pollefeys, and R.~Urtasun.
\newblock Box in the box: Joint 3d layout and object reasoning from single
  images.
\newblock In {\em ICCV 2013}.

\bibitem{schwing2012efficient}
A.~G. Schwing, T.~Hazan, M.~Pollefeys, and R.~Urtasun.
\newblock Efficient structured prediction for 3d indoor scene understanding.
\newblock In {\em CVPR 2012}.

\bibitem{song2016deep}
S.~Song and J.~Xiao.
\newblock Deep sliding shapes for amodal 3d object detection in rgb-d images.
\newblock In {\em CVPR 2016}.

\bibitem{song2014sliding}
S.~Song and J.~Xiao.
\newblock Sliding shapes for 3d object detection in depth images.
\newblock In {\em ECCV 2014}.

\bibitem{van2008visualizing}
L.~Van~der Maaten and G.~Hinton.
\newblock Visualizing data using t-sne.
\newblock {\em Journal of Machine Learning Research 2008}.

\bibitem{wang2010discriminative}
H.~Wang, S.~Gould, and D.~Koller.
\newblock Discriminative learning with latent variables for cluttered indoor
  scene understanding.
\newblock {\em ECCV 2010}.

\bibitem{yu2015multi}
F.~Yu and V.~Koltun.
\newblock Multi-scale context aggregation by dilated convolutions.
\newblock {\em ICLR 2016}.

\bibitem{ren2016cfile}
R.~Yuzhuo, C.~Chen, L.~Shangwen, and K.~C.-C.~Jay.
\newblock A coarse-to-fine indoor layout estimation (cfile) method.
\newblock In {\em ACCV 2016}.

\bibitem{zhang2010supervised}
H.~Zhang, J.~Xiao, and L.~Quan.
\newblock Supervised label transfer for semantic segmentation of street scenes.
\newblock In {\em ECCV 2010}.

\bibitem{zhao2013scene}
Y.~Zhao and S.-C. Zhu.
\newblock Scene parsing by integrating function, geometry and appearance
  models.
\newblock In {\em CVPR 2013}.

\end{thebibliography}
}

\end{document}